# Fuzzy Mixed Integer Linear Programming for Air Vehicles Operations Optimization

Arindam Chaudhari[1], Dipak Chatterjee[2] and Ritesh Rajput[3]

[1]Assistant Professor (Computer Science Engineering),
NIIT University, Neemrana, Rajasthan, India
arindam_chau@yahoo.co.in
[2]Principal and Professor (Mathematics),
Institute of Engineering and Management, West Bengal, Kolkata, India
[3]BTech (Computer Science Engineering) 3rd Year,
NIIT University, Neemrana, Rajasthan, India

**Abstract.** Multiple Air Vehicles (AVs) to prosecute geographically dispersed targets is an important optimization problem. Associated multiple tasks viz., target classification, attack and verification are successively performed on each target. The optimal minimum time performance of these tasks requires cooperation among vehicles such that critical time constraints are satisfied i.e. target must be classified before it can be attacked and AV is sent to target area to verify its destruction after target has been attacked. Here, optimal task scheduling problem from Indian Air Force is formulated as Fuzzy Mixed Integer Linear Programming (FMILP) problem. The solution assigns all tasks to vehicles and performs scheduling in an optimal manner including scheduled staged departure times. Coupled tasks involving time and task order constraints are addressed. When AVs have sufficient endurance, existence of optimal solution is guaranteed. The solution developed can serve as an effective heuristic for different categories of AV optimization problems.

**Keywords:** Fuzzy Mixed Integer Linear Programming, Air Vehicles, Planning, Scheduling

## 1 Introduction

Unmanned Air Vehicle (UAV) is an aircraft with no onboard pilot which can be remote controlled or fly autonomously based on pre-programmed flight plans with complex dynamic automation systems. Optimizing air to ground operations of such air vehicles is an important decision making problem [7]. A more challenging scenario is multiple UAVs required to service geographically dispersed targets. Multiple tasks are required to be successively performed on each target, viz. targets to be classified, attacked and damage inflicted on targets must be assessed. The floating time constraints are critical. A target cannot be attacked before it is classified and UAV sent to target area only after target attack has been executed. Multi role UAVs are considered such that each UAV can perform all tasks. Small UAVs such as autonomous Wide Area Search Munitions (WASM) are deployed in groups from aircraft flying at higher altitudes. They are individually capable of autonomous searching, recognizing and attacking targets. The ability to network involving communication of target information to one another and consequent co-operation greatly improves effectiveness of future UAV teams. The problem comprises of planning performance of UAVs' tasks such that critical timing constraints are satisfied. This calls for optimal assignment and scheduling [1].

In a time phased network optimization model was used to perform task allocation for team of UAVs [9]. The model is run simultaneously on all air vehicles at discrete points in time and assigns each vehicle one or more tasks each time it is executed. The network optimization model is iteratively implemented such that all known targets are prosecuted by resulting allocation. The model is solved each time new information is brought in the system, because a new target has been discovered or known target's status is changed, thus achieving feedback action. Classification, attack and battle damage assessment tasks can all be assigned to different vehicles when a target is found resulting in target being more quickly serviced. A single vehicle can be given multiple task assignments to be performed in succession if that is more efficient than having multiple vehicles perform tasks individually. In [9], variable path lengths are included to guarantee that feasible trajectories calculated for all tasks. This method is computationally efficient and scales well. However, the iterative procedure is heuristic and suboptimal. Tabu Search can be used to solve difficult combinatorial optimization problems such as vehicle routing problem with fixed time windows [2], [3], [4], [5], [11].

In this work, a solution method for AV routing combinatorial optimization problem with floating time windows from Indian Air Force is developed. The work addresses optimal formulation for solving coupled multiple-assignment and scheduling problem. Time variables used are continuous in nature. They are modeled using fuzzy membership functions to tackle the inherent uncertainty involved. This leads to the formulation of AV optimization problem as Fuzzy Mixed Integer Linear Programming (FMILP) problem [8], [12] and allows optimal solution to be obtained while satisfying all time constraints. The entire FMILP problem formulation is developed using MATLAB. The mathematical formulation presented can be used to solve optimally for some realistic problem sizes. The method presented here allows staged departure times before vehicles begin their tour of targets and tasks. FMILP formulation is flexible enough to allow consideration of different cost functions such as, mission completion in minimum time, shortest total path lengths travelled by vehicles or maximization of number of AV that survive the mission. The method also accommodates fixed time windows as in vehicle routing problem [2], [5], [11]. FMILP formulation also incorporates dynamic and logical constraints on task performance as in scheduling problems [7]. Without using fixed time windows, feasibility is guaranteed as long as number of air vehicles exceeds number of targets, even with three or more tasks per target provided air vehicles have sufficient endurance.

This paper is organized as follows. In section 2, UAV optimization problem is given. This is followed by mathematical formulation of AV optimization problem in section 3. In the next section, experimental results are given. In section 5, a discussion on problem size is illustrated. Finally, in section 6 conclusions are given.

## 2 Unmanned Air Vehicles Optimization Problem

In this section AV optimization problem scenario is presented which is adapted from Indian Air Force. Consider $n$ geographically dispersed targets with known position $w$ ($w \geq n+1$) air vehicles (AV). We then have $n+w+1$ nodes: $n$ target nodes, $w$ source nodes (points of origins of AVs) and one sink node. Nodes $1,......,n$ are located at $n$ target positions. Nodes $n+1,........,n+w$ are located at vehicle initial positions. Node $n+w+1$ is sink. An AV with no future target assignments is relegated to sink i.e. will continue to search. A vehicle located at sink cannot be reassigned. The flight time of AV $v$ from node $i$ to node $j$ to perform task $k$ at node $j$ is:

$$\tilde{t}_{ij}^{(v,k)} \geq 0; i = 1,......., n+w; j = 1,......., n; v = 1,......., w \quad (1)$$

which is modeled as the following fuzzy membership function [10]:

$$\tilde{t}_{ij}^{(v,k)} = \begin{cases} 1+(z_j - c_{ij})/b_{ij}^- , & -b_{ij}^- \leq (z_j - c_{ij}) \leq 0 \\ 1-(z_j - c_{ij})/b_{ij}^+ , & 0 \leq (z_j - c_{ij}) \leq b_{ij}^+ \\ 0 & otherwise \end{cases} \quad (2)$$

Here, $z_j$ is $j^{th}$ input of $i^{th}$ fuzzy membership function, $c_{ij}$ is modal point, $b_{ij}^-$ its lower half width and $b_{ij}^+$ upper half width. The membership function attains value 1 when input is $c_{ij}$. As input decreases from $c_{ij}$ membership function value decreases to 0 at $c_{ij} - b_{ij}^-$ and remains at 0 for all inputs less than $c_{ij} - b_{ij}^-$. As input increases from $c_{ij}$, membership function value decreases to 0 at $c_{ij} + b_{ij}^+$ and remains at 0 for all inputs greater than $c_{ij} + b_{ij}^+$. The obvious choice of this membership function arises because there is only one output here viz. time.

The time to travel from node $i$ to node $j$ depends on particular AV $v's$ airspeed and assigned task $k$. Three tasks must be performed on each target viz. (a) classification (b) attack (c) target damage assessment (verification). Furthermore, once an AV attacks a target, it is destroyed and can no longer perform additional tasks. This is certainly the case for powered munitions and WASM mission, but if AV was a reusable aircraft, the problem formulation would have to be modified and account for AV's depletion of its store of ammunition following each attack. Three tasks must be performed on each target in order listed. This results in critical timing constraints. The problem inherits some aspects of job shop scheduling [7]. In optimization problem considered, the number of pre-specified problem parameters is $3wn + 3n(n-1)w = 3n^2w$. When Euclidean distances are used the dimension of parameter space is reduced appreciably. The endurance of AV $v$ is $T_v, v = 1,......., w$.

## 3 Mathematical formulation of Air Vehicles Optimization Problem

In this section, the mathematical formulation of AV optimization problem given in section 2 is illustrated. The problem is modeled using FMILP technique. The FMILP model uses a discrete approximation of real world based on nodes that represent start and end positions for segments of UAVs path. Nodes representing target positions range from $1,.......n$ and nodes for initial UAV positions range from $n+1,......., n+w$. There is also an additional node for sink $n+w+1$. The sink node is used when no further assignment for UAV is present. It goes to sink when it has completed all its tasks or when not assigned to another task. Generally, when a UAV enters sink it is used for performing search of battle space. FMILP model requires information on costs or times for UAV to fly from one node to another node. The flight times are represented by $\tilde{t}_{ij}^{(v,k)} \geq 0$ as given in Equation (2), the time it takes UAV $v$ to fly from node $i$ to node $j$ to perform task $k$.

## 3.1 Decision Variables

The binary decision variable $x_{ij}^{(v,k)} = 1$ AV $v$ is assigned to fly from node $i$ to node $j$ to perform task $k$ at node $j$ and $0$ otherwise; $i = 1,\ldots,n+w$; $j = 1,\ldots,n$; $v = 1,\ldots,w$; $k = 1,2,3$. For $i = n+1,\ldots,n+w$, only $x_{n+v,j}^{(v,k)}$ exist. These variables correspond with first task assignment each vehicle receives starting from its unique source node. Only vehicle $v$ can do a task starting from node $n+v$. For task assignments $k = 1,3, i \neq j$ and for task assignment $k = 2, i = j$ is allowed. The later allows for an AV to perform target classification task and immediately thereafter attack the target. Thus, there are $wn(3n+1)$ binary decision variables. Another decision variable $x_{i,n+w+1}^{(v,k)} = 1$ if AV $v$ is assigned to fly node $i$ to sink node $n+w+1$ and $0$ otherwise $v = 1,\ldots,w$; $i = 1,\ldots,n+w$. This adds $w(n+1)$ binary decision variables. Entering sink can also be thought of as being reassigned to search task. Time of performance of task $k$ on target $j$ is $\tilde{t}_j^{(k)} > 0$; $k = 1,2,3$; $j = 1,\ldots,n$ which is modeled as the following fuzzy membership function [6]:

$$\tilde{t}_j^{(k)} = \begin{cases} 0 & , x = a \\ x/(x_i - a) + a/(a - x_i) & , a \leq x \leq x_i \\ 0 & x > b \end{cases} \quad (3)$$

These are continuous decision variables which are $3n$ in number. There are also $w$ additional continuous decision variables: time AV $v$ leaves node $j = n+v$ is $\tilde{t}_v$; $v = 1,\ldots,w$ which is also modeled as fuzzy membership function given by Equation (3). There are $w[n(3n+2)+1]$ and $3n+w$ binary and continuous non–negative decision variables.

## 3.2 Cost Functions

The cost functions involved for AV optimization problem include the following:

(a) Minimize the total flight time of the AVs: $J = \sum_{k=1}^{3}\sum_{v=1}^{w}\sum_{i=1}^{n+w}\sum_{j=1}^{n} \tilde{t}_{i,j}^{(v,k)} x_{i,j}^{(v,k)}$ (4)

(b) Alternatively, minimize total engagement time. Target $j$ is visited for last time at time $\tilde{t}_j^{(3)}$. Let $\tilde{t}_f$ be the time at which all targets have been through verification. Considering an additional continuous decision variable $\tilde{t}_f \in \mathcal{R}_+^1$. The cost function is then $J = tf$ and we minimize $J$ subject to constraints

$$\tilde{t}_j^{(3)} \leq \tilde{t}_f; j = 1,\ldots,n \quad (5)$$

A small weightage to time of performance of each individual task is added to encourage each task to be completed as quickly as possible. Then,

$$J = \tilde{t}_f + \tilde{c}_j^{(k)}\tilde{t}_j^{(k)}; j = 1,\ldots,n; k = 1,2,3 \quad (6)$$

where, $\tilde{c}_j^{(k)} > 0$ is small weight on completion time of each individual task and modeled as fuzzy membership function given by Equation (3).

(c) The number $n_s$ of surviving UAVs that end up in sink and the corresponding cost function is given by,

$$J = n_s = \sum_{v=1}^{w} \sum_{i=1}^{n+w} x_{i,n+w+1}^{(v)} \quad (7)$$

and optimization problem considered is $\max J = n_s$.

### 3.3 Constraints

The corresponding constraints for AV optimization problem require inclusion of all issues which are critical to automatically enforcing desired vehicle behavior.

(a) Mission completion constraints: It requires that all three tasks are performed on each target exactly one time. The following must hold as evident for linear assignment problems [1], [7] yielding $3n$ constraints.

$$\sum_{v=1}^{w} \sum_{i=1, i \neq j}^{n+w} x_{ij}^{(v,k)} = 1 \ k = 1,3; \ j = 1,\ldots,n \quad (8) \qquad \sum_{v=1}^{w} \sum_{i=1, i \neq j}^{n+w} x_{ij}^{(v,k)} = 1 \ j = 1,\ldots,n \quad (9)$$

(b) Not more than one AV is assigned to perform a specific task $k$ on a specified target $j$. This also yields $3n$ constraints.

(c)
$$\sum_{i=1, i \neq j}^{n+w} x_{i,j}^{(v,k)} \leq 1 \ v = 1,\ldots,w; \ j = 1,\ldots,n; k = 1,3 \quad (10)$$

$$\sum_{i=1}^{n+w} x_{i,j}^{(v,2)} \leq 1 \ v = 1,\ldots,w; \ j = 1,\ldots,n \quad (11)$$

(c) AV $v$ coming from outside can visit target $j$ at most once.

$$\sum_{k=1}^{3} \sum_{i=1, i \neq j}^{n+w} x_{i,j}^{(v,k)} \leq 1 \ v = 1,\ldots,w; \ j = 1,\ldots,n \quad (12)$$

In addition each AV $v$ can only enter the sink once. This yields $w(n+1)$ constraints.

$$\sum_{i=1}^{n+w} x_{i,n+w+1}^{(v)} \leq 1 \ v = 1,\ldots,w \quad (13)$$

(d) An AV $v$ leaves node $j$ at most once yielding $nw$ constraints.

$$\sum_{k=1}^{3} \sum_{i=1, i \neq j}^{n} x_{i,j}^{(v,k)} + x_{j,n+w+1}^{(v)} \leq 1 \ v = 1,\ldots,w; \ j = 1,\ldots,n \quad (14)$$

The constraints (c) and (d) eliminate the possibility of loops.

(e) Perishable Munition: An AV $v$ can be assigned to attack at most one target yielding $w$ constraints.

$$\sum_{j=1}^{n}\sum_{i=1}^{n+w} x_{i,j}^{(v,2)} \leq 1 \quad \forall v = 1,\ldots, w \quad (15)$$

(f) If AV $v$ is assigned to fly to target $j$ for verification. It cannot possibly be assigned to attack target $j$.

$$\sum_{i=1,i\neq j}^{n+w} x_{i,j}^{(v,2)} \leq 1 - \sum_{i=1,i\neq j}^{n+w} x_{i,j}^{(v,3)} \quad v = 1,\ldots, w; j = 1,\ldots, n \quad (16)$$

(g) Continuity Constraints: These constraints ensure proper flow balance is maintained at each node.

(i) If AV $v$ enters target node $j$ for purpose of performing task 3 it must also exit target node $j$.

$$\sum_{i=1,i\neq j}^{n+w} x_{i,j}^{(v,3)} \leq \sum_{k=1}^{3}\sum_{i=1,i\neq j}^{n} x_{j,i}^{(v,k)} + x_{j,n+w+1}^{(v)} \quad v = 1,\ldots, w; j = 1,\ldots, n \quad (17)$$

(ii) If AV $v$ enters target node $j$ for purpose of performing task 1 it must also exit target node $j$.

$$\sum_{i=1,i\neq j}^{n+w} x_{i,j}^{(v,1)} \leq \sum_{k=1}^{3}\sum_{i=1,i\neq j}^{n} x_{j,i}^{(v,k)} + x_{j,j}^{(v,2)} + x_{j,n+w+1}^{(v)} \quad v = 1,\ldots, w; j = 1,\ldots, n \quad (18)$$

(iii) As a munition is perishable if AV $v$ is assigned to fly to target node $j$ to perform task $k = 2$ then at any other point in time, AV $v$ cannot be assigned to fly from target $j$ to target $i, i \neq j$ to perform any other task $k$ at target $i$. AV $v$ can enter target $j$ not more than once.

$$\sum_{k=1}^{3}\sum_{i=1,i\neq j}^{n} x_{j,i}^{(v,k)} + x_{j,n+w+1}^{(v)} \leq 1 - \sum_{i=1}^{n+w} x_{i,j}^{(v,2)} \quad v = 1,\ldots, w; j = 1,\ldots, n \quad (19)$$

(iv) If AV $v$ is not assigned to visit node $j$, then it cannot be assigned to fly out of node $j$.

$$\sum_{k=1}^{3}\sum_{i=1,i\neq j}^{n} x_{j,i}^{(v,k)} + x_{j,n+w+1}^{(v)} \leq \sum_{k=1}^{3}\sum_{i=1,i\neq j}^{n+w} x_{i,j}^{(v,k)} \quad v = 1,\ldots, w; j = 1,\ldots, n \quad (20)$$

(v) All AVs leave source nodes. An AV leaves source node even if this entails direct assignment to sink.

$$\sum_{k=1}^{3}\sum_{j=1}^{n} x_{n+v,j}^{(v,k)} + x_{n+v,n+w+1}^{(v)} = 1 \quad \forall v = 1,\ldots, w \quad (21)$$

(vi) An AV cannot attack target node $i$ coming from target node $i$ unless it entered target node $i$ to perform classification.

$$x_{i,i}^{(v,2)} \leq \sum_{j=1}^{n+w} x_{j,i}^{(v,1)} \quad \forall i = 1,\ldots, n; v = 1,\ldots, w \quad (22)$$

(h) Timing constraints: Nonlinear equations enforcing timing constraints can be formulated represented as FMILP. Thus, assuming

$$T \equiv \max_{v}\{\tilde{T}_v\}_{v=1}^{w} \quad (23)$$

Then the linear timing constraints take the following form:

$$\tilde{t}_j^{(k)} \leq \tilde{t}_i^{(1)} + \tilde{t}_{i,j}^{(v,k)} + \left(2 - x_{i,j}^{(v,k)} - \sum_{l=1,l\neq i}^{n+w} x_{l,i}^{(v,1)}\right)wT \quad (24)$$

$$\tilde{t}_j^{(k)} \geq \tilde{t}_i^{(1)} + \tilde{t}_{i,j}^{(v,k)} - \left(2 - x_{i,j}^{(v,k)} - \sum_{l=1, l \neq i}^{n+w} x_{l,i}^{(v,1)}\right)wT \quad (25)$$

$$\tilde{t}_j^{(k)} \leq \tilde{t}_i^{(3)} + \tilde{t}_{i,j}^{(v,k)} + \left(2 - x_{i,j}^{(v,k)} - \sum_{l=1, l \neq i}^{n+w} x_{l,i}^{(v,3)}\right)wT \quad (26)$$

$$\tilde{t}_j^{(k)} \geq \tilde{t}_i^{(3)} + \tilde{t}_{i,j}^{(v,k)} - \left(2 - x_{i,j}^{(v,k)} - \sum_{l=1, l \neq i}^{n+w} x_{l,i}^{(v,3)}\right)wT \quad (27)$$

$$\forall i = 1,......., n; j = 1,......., n; i \neq j; v = 1,......., w; k = 1,3$$

In addition to this we also have the following timing constraints:

$$\tilde{t}_j^{(2)} \leq \tilde{t}_i^{(1)} + \tilde{t}_{i,j}^{(v,2)} + \left(2 - x_{i,j}^{(v,2)} - \sum_{l=1, l \neq i}^{n+w} x_{l,i}^{(v,1)}\right)wT \quad (28)$$

$$\tilde{t}_j^{(2)} \geq \tilde{t}_i^{(1)} + \tilde{t}_{i,j}^{(v,2)} - \left(2 - x_{i,j}^{(v,2)} - \sum_{l=1, l \neq i}^{n+w} x_{l,i}^{(v,1)}\right)wT \quad (29)$$

$$\tilde{t}_j^{(2)} \leq \tilde{t}_i^{(3)} + \tilde{t}_{i,j}^{(v,2)} + \left(2 - x_{i,j}^{(v,2)} - \sum_{l=1, l \neq i}^{n+w} x_{l,i}^{(v,3)}\right)wT \quad (30)$$

$$\tilde{t}_j^{(2)} \geq \tilde{t}_i^{(3)} + \tilde{t}_{i,j}^{(v,2)} - \left(2 - x_{i,j}^{(v,2)} - \sum_{l=1, l \neq i}^{n+w} x_{l,i}^{(v,3)}\right)wT \quad (31)$$

$$\forall j = 1,......., n; k = 1,2,3; v = 1,......., w$$

*Also*

$$\tilde{t}_j^{(k)} \leq \tilde{t}_v + \tilde{t}_{n+v,j}^{(v,k)} + \left(1 - x_{n+v,j}^{(v,k)}\right)wT \quad (32)$$

$$\tilde{t}_j^{(k)} \leq \tilde{t}_v + \tilde{t}_{n+v,j}^{(v,k)} - \left(1 - x_{n+v,j}^{(v,k)}\right)wT \quad (33)$$

$$\forall j = 1,......., n; k = 1,2,3; v = 1,......., w$$

These timing constraints operate in pairs. The inequalities (32) – (33) are not validated for assignments $x_{i,j}^{(v,k)}$ that do not occur but become soft equality constraint for assignments that occur. Thus, the time that a task $k$ is performed on target $j$ by AV $v$ will be equal to the time that preceding task was performed by AV $v$ at node $i$ plus the time it will take AV $v$ to fly from node $i$ to node $j$. A similar constraint applies if AV $v$ left its source node $n + v$ to fly to node $j$. Also

$$t_j^{(1)} \leq t_j^{(2)} < t_j^{(3)}; j = 1,......., n \quad (34)$$

The timing constraints add $2n [w (6n – 1) + 1]$ linear inequality constraints. The constraints (24) – (34) are critical for FMILP formulation of AV optimization problem.

## 4 Experimental Results

In this section one real life example from Indian Air Force comprising of three UAVs and one Target to illustrate the mathematical formulation of AV optimization problem presented in section 3.

## 4.1 Three UAVs and One Target

We consider the case of three AVs and one target i.e. $w = 3$ and $n = 1$. Though the number of variables in the problem is small enough, this can easily be extended to large number of variables at the cost of additional computational complexity. There are 18 binary and 6 continuous decision variables which are given in equations below. Minimizing time the final task occurs will add an additional continuous decision variable for a total of 25 decision variables.

$$(x_1,........, x_5) = (x_{1,1}^{(1,2)}, x_{1,1}^{(2,2)}, x_{1,1}^{(3,2)}, x_{2,1}^{(1,1)}, x_{2,1}^{(1,2)}) \quad (x_6,........, x_{10}) = (x_{2,1}^{(1,3)}, x_{3,1}^{(2,1)}, x_{3,1}^{(2,2)}, x_{3,1}^{(2,3)}, x_{4,1}^{(3,1)})$$

$$(x_{11},........, x_{15}) = (x_{4,1}^{(3,2)}, x_{4,1}^{(3,3)}, x_{1,5}^{(1)}, x_{1,5}^{(2)}, x_{1,5}^{(3)}) \quad (x_{16}, x_{17}, x_{18}) = (x_{2,5}^{(1)}, x_{3,5}^{(2)}, x_{4,6}^{(3)}) \quad (35)$$

$$(x_{19},........, x_{25}) = (\tilde{t}_1, \tilde{t}_2, \tilde{t}_{3..}, \tilde{t}_1^{(1)}, \tilde{t}_1^{(2)}, \tilde{t}_1^{(3)}, \tilde{t}) \quad (36)$$

We wish to minimize

$$J = x_{25} + 0.1(x_{22} + x_{23} + x_{24}) \quad (37)$$

subject to following constraints:

Constraint 1:

$$x_4 + x_7 + x_{10} = 1 \qquad x_6 + x_9 + x_{12} = 1 \qquad x_1 + x_2 + x_3 + x_5 + x_8 + x_{11} = 1 \quad (38)$$

Constraint 2:

$$x_4 + x_7 + x_{10} \leq 1 \qquad x_6 + x_9 + x_{12} \leq 1 \qquad x_1 + x_2 + x_3 + x_5 + x_8 + x_{11} \leq 1 \quad (39)$$

Constraint 3:

$$x_4 + x_5 + x_6 \leq 1 \qquad x_7 + x_8 + x_9 \leq 1 \qquad x_{10} + x_{11} + x_{12} \leq 1 \quad (40)$$

Constraint 7(a):

$$x_6 \leq x_{13} \qquad x_9 \leq x_{14} \qquad x_{12} \leq x_{15} \quad (41)$$

Constraint 7(b):

$$x_4 \leq x_1 + x_{13} \qquad x_7 \leq x_2 + x_{14} \qquad x_{10} \leq x_3 + x_{15} \quad (42)$$

Constraint 7(c):

$$x_1 + x_5 + x_{13} \leq 1 \qquad x_2 + x_8 + x_{14} \leq 1 \qquad x_3 + x_{11} + x_{15} \leq 1 \quad (43)$$

Constraint 7(d):

$$x_{13} \leq x_4 + x_6 \qquad x_{14} \leq x_7 + x_9 \qquad x_{15} \leq x_{10} + x_{12} \quad (44)$$

Constraint 7(e):

$$x_4 + x_5 + x_6 + x_{16} = 1 \qquad x_7 + x_8 + x_9 + x_{17} = 1 \qquad x_{10} + x_{11} + x_{12} + x_{18} = 1 \quad (45)$$

Constraint 7(f):

$$x_1 \leq x_4 \qquad x_2 \leq x_7 \qquad x_3 \leq x_{10} \quad (46)$$

It is to be noted that constraints (d)–(f) are eliminated in one-target case. With only one target, constraints associated with Equations (24)–(27), (30) and (31) are ineffective, which results in following timing constraints. From Equations (28) and (29),

$$x_{23} \leq x_{22} + \tilde{t}_{1,1}^{(1,2)} + (2 - x_1 - x_4)wT \qquad x_{23} \geq x_{22} + \tilde{t}_{1,1}^{(1,2)} - (2 - x_1 - x_4)wT$$
$$x_{23} \leq x_{22} + \tilde{t}_{1,1}^{(2,2)} + (2 - x_2 - x_7)wT \qquad x_{23} \geq x_{22} + \tilde{t}_{1,1}^{(2,2)} - (2 - x_2 - x_7)wT \quad (47)$$
$$x_{23} \leq x_{22} + \tilde{t}_{1,1}^{(3,2)} + (2 - x_3 - x_{10})wT \qquad x_{23} \geq x_{22} + \tilde{t}_{1,1}^{(3,2)} - (2 - x_3 - x_{10})wT$$

From Equations (32) and (33),
$$x_{22} \leq x_{19} + \tilde{t}_{2,1}^{(1,1)} + (1 - x_4)wT \qquad x_{22} \geq x_{19} + \tilde{t}_{2,1}^{(1,1)} - (1 - x_4)wT$$
$$x_{22} \leq x_{20} + \tilde{t}_{3,1}^{(2,1)} + (1 - x_7)wT \qquad x_{22} \geq x_{20} + \tilde{t}_{3,1}^{(2,1)} - (1 - x_7)wT \quad (48)$$
$$x_{22} \leq x_{21} + \tilde{t}_{4,1}^{(3,1)} + (1 - x_{10})wT \qquad x_{22} \geq x_{21} + \tilde{t}_{4,1}^{(3,1)} - (1 - x_{10})wT$$

and
$$x_{23} \leq x_{19} + \tilde{t}_{2,1}^{(1,2)} + (1 - x_5)wT \qquad x_{23} \geq x_{19} + \tilde{t}_{2,1}^{(1,2)} - (1 - x_5)wT$$
$$x_{23} \leq x_{20} + \tilde{t}_{3,1}^{(2,2)} + (1 - x_8)wT \qquad x_{23} \geq x_{20} + \tilde{t}_{3,1}^{(2,1)} - (1 - x_8)wT \quad (49)$$
$$x_{23} \leq x_{21} + \tilde{t}_{4,1}^{(3,2)} + (1 - x_{11})wT \qquad x_{22} \geq x_{21} + \tilde{t}_{4,1}^{(3,2)} - (1 - x_{11})wT$$

and finally
$$x_{24} \leq x_{19} + \tilde{t}_{2,1}^{(1,3)} + (1 - x_6)wT \qquad x_{24} \geq x_{19} + \tilde{t}_{2,1}^{(1,3)} - (1 - x_6)wT$$
$$x_{24} \leq x_{20} + \tilde{t}_{3,1}^{(2,3)} + (1 - x_9)wT \qquad x_{24} \geq x_{20} + \tilde{t}_{3,1}^{(2,3)} - (1 - x_9)wT \quad (50)$$
$$x_{24} \leq x_{21} + \tilde{t}_{4,1}^{(3,3)} + (1 - x_{12})wT \qquad x_{24} \geq x_{21} + \tilde{t}_{4,1}^{(3,3)} - (1 - x_{12})wT$$

Also from Equation (34), we have
$$x_{22} \leq x_{23} - \varepsilon \qquad (51) \qquad\qquad x_{23} \leq x_{24} - \varepsilon \qquad (52)$$
where, $\varepsilon > 0$ is small positive constant which is modeled using fuzzy membership function. Considering $\varepsilon = 0.1$ we enforce a small delay between each task being performed on a target. Finally, from Equation (5) we have,
$$x_{24} \leq x_{25} \qquad (53)$$

Thus, full set of constraints contain 6 constraints, 51 inequality constraints for 57 total constraints. Assuming a simplifying situation that time to travel from node $i$ to $j$ to perform task $k$ is independent of which task is required and vehicle performing the task. Then, $\tilde{t}_{i,j}^{(v,k)} = \tilde{t}_{i,j}$. For example,
$$t_{1,1} = 0.11 \qquad t_{2,1} = 3.8 \qquad t_{3,1} = 4.24 \qquad t_{4,1} = 5.38$$
We set $T = 100$ as endurance of all AVs, so that endurance is not a constraint and feasibility is guaranteed. Then optimal assignment is as follows:
$$x_i = 1; \qquad\qquad i = 1,4,10,14,18$$
$$x_i = 0; \qquad i = 2,3,5,....,8,9,....,13,15,16,17$$
$$x_i = 0; \qquad\qquad i = 19,20,21$$
$$x_{22} = 3.7 \qquad x_{23} = 3.81 \qquad x_{24} = 4.24 \qquad x_{25} = 4.24$$

This corresponds with all three vehicles immediately leaving their source nodes $(x_{19} - x_{21} = 0)$ and vehicle 1 performing classification and attack on target at $t = 3.7$ and $t = 3.81$ respectively with vehicle 2 performing verification at $T = 4.24$. Vehicle 3 flies in direction to sink. Suppose it takes longer for a vehicle that has just classified a target to complete an attack on that target. Then we have the initial conditions as:

$$t_{1,1} = 1.1 \qquad t_{2,1} = 3.8 \qquad t_{3,1} = 4.24 \qquad t_{4,1} = 5.38$$

In this case, assignment is identical, except that attack occurs at $t = 4.64$ and verification at $t = 4.69$. This example illustrates the situation where verification had to be delayed such that it took place after attack. Finally, suppose that vehicle 3 is closer to target initially, the initial conditions are:

$$t_{1,1} = 1.1 \qquad t_{2,1} = 3.8 \qquad t_{3,1} = 4.24 \qquad t_{4,1} = 4.50$$

Then optimal assignment is as follows:

$$x_i = 1; \qquad\qquad i = 4,7,12,13,15$$
$$x_i = 0; \quad i = 1,2,3,5,6,8,9,10,11,14,16,17,18,19$$
$$x_i = 0; \qquad\qquad i = 19,20,21$$
$$x_{22} = 3.7 \qquad x_{23} = 4.24 \qquad x_{24} = 4.50 \qquad x_{25} = 4.50$$

This assignment requires all three vehicles to immediately leave their source nodes and proceed to target. Vehicle 1 performed classification, vehicle 2 attack and vehicle 3 verification tasks. WASMs 1 and 3 then proceed to sink i.e., continue to search for other targets.

## 5 Discussion on Problem size

Considering the AV optimization problem in section 3 and corresponding illustrative example in section 4, a discussion on problem size is presented here. For $n$ targets, $w$ vehicles and $m = 3$ tasks per target, problem size comprises of $n(n-1)wm + nwm + 2nw + mn + 2w + 1$ decision variables. Of these $3 + nm + 1$ are continuous timing variables and rest is binary decision variables. The number of constraints also grows rapidly. There are $12(n-1)nw + 9nw + 2nwm + 2nm + 3w$ constraints, of which mn + w are equality constraints and $7nw + 2w$ inequality non-timing constraints and $12(n-1)nw + 2nw + 2nwm + mn$ inequality timing constraints. The size of FMILP increases rapidly as problem size increases. Some practical sized problems are amenable to optimal solution with this formulation. For $n = 2$, $w = 3$, there are 51 binary decision variables, 10 continuous decision variables, 9 linear equality constraints and 174 linear inequality constraints. For $n = 2$, $w = 4$, there are 68 binary decision variables, 11 continuous decision variables, 10 linear equality constraints and 230 linear inequality constraints. For $n = 2$, $w = 5$, there are 85 binary decision variables, 12 continuous decision variables, 11 linear equality constraints and 286 linear inequality constraints. For $n = 3$, $w = 4$, there are 136 binary decision variables, 14 continuous decision variables, 13 linear equality constraints and 485 linear inequality constraints. The growth of constraints and variables is linear in number of vehicles but quadratic in number of targets. For many operational scenarios involving large problem sizes optimal solutions cannot be found within available computation time and the problem becomes intractable in nature leading to NP Complete problem.

# 6 Conclusion

In this work, FMILP model is presented to find an optimal solution to NP Complete complex multiple task assignment and scheduling problem where tasks are coupled by timing and task precedence constraints. With due consideration to linearity and inherent uncertainty in real life data involved, AV optimization problem is formulated as FMILP by treating the time variable as continuous one. The mathematical formulation allows staged AV departure times to guarantee that time constraints are satisfied and incorporate varying task completion times into optimization model. Here feasibility is guaranteed provided that number of UAVs employed exceeds number of targets and UAV endurance is sufficiently high. The formulation is flexible enough to allow for various alternative performance functionals. The rigorous formulation allows true optimal solution for challenging assignment and scheduling problem. Experimental results are presented for practical real life problem from Indian Air Force which demonstrates the efficiency of FMILP problem formulation. The solution developed can serve as an effective heuristic for different categories of AV optimization problems.